# Flood Prediction Using Machine Learning Models

Miah Mohammad Asif Syeed
*Department of Computer Science and Engineering*
*BRAC University*
Dhaka, Bangladesh
miah.mohammad.asif.syeed@g.bracu.ac.bd

Maisha Farzana
*Department of Computer Science and Engineering*
*BRAC University*
Dhaka, Bangladesh
maisha.farzana1@g.bracu.ac.bd

Ishadie Namir
*Department of Computer Science and Engineering*
*BRAC University*
Dhaka, Bangladesh
ishadie.namir@g.bracu.ac.bd

Ipshita Ishrar
*Department of Computer Science and Engineering*
*BRAC University*
Dhaka, Bangladesh
ipshita.ishrar@g.bracu.ac.bd

Meherin Hossain Nushra
*Department of Computer Science and Engineering*
*BRAC University*
Dhaka, Bangladesh
meherin.hossain.nushra@g.bracu.ac.bd

Tanvir Rahman
*Department of Computer and Information Sciences*
*School of Engineering*
*University of Delaware*
rtanvir@udel.edu

*Abstract*- Floods are one of nature's most catastrophic calamities which cause irreversible and immense damage to human life, agriculture, infrastructure and socio-economic system. Several studies on flood catastrophe management and flood forecasting systems have been conducted. The accurate prediction of the onset and progression of floods in real time is challenging. To estimate water levels and velocities across a large area, it is necessary to combine data with computationally demanding flood propagation models. This paper aims to reduce the extreme risks of this natural disaster and also contributes to policy suggestions by providing a prediction for floods using different machine learning models. This research will use Binary Logistic Regression, K-Nearest Neighbor (KNN), Support Vector Classifier (SVC) and Decision tree Classifier to provide an accurate prediction. With the outcome, a comparative analysis will be conducted to understand which model delivers a better accuracy.

## I. INTRODUCTION

Flood is caused by an overflow of water from a lake, river, or ocean that submerges neighboring land. Every year, floods affect roughly 4.84 million people in India, 3.84 million in Bangladesh, and 3.28 million in China [1]. Other countries' cities are also prone to flooding. The Netherlands, Monaco, Bahrain, and other low-lying areas are at risk of floods. Australia's floods claimed 73 lives between 1997 and 2008 [2]. Floods in the United States kill roughly 100 people and cost $7.5 billion in damage yearly [3]. According to the World Resource Institute, floods would affect 147 million people by 2030, causing $174 billion to $712 billion in property damage [4].

Bangladesh is one of the most vulnerable countries in South Asia to natural disasters, especially floods. According to a study [5], over 80% of Bangladesh's land is in flood plains. According to the study [5], Bangladesh was hit by floods 78 times between 1971 and 2014, killing 41,783 people.

According to a study, the NARX structure of the Artificial Neural Network model [6] can predict floods 5 hours in advance with 73.54% accuracy. This study used upstream river water levels as an input variable because they influence flooding. Another study [7] used Back Propagation Neural Networks (BPNN) to forecast flood levels (BPN). The BPN model uses water level data from stations in Johor, Malaysia. The model's output was unsatisfactory, so an Extended Kalman Filter (EKF) was added to improve it. The paper claims that adding EKF to the BPN model resulted in better accuracy. The EKF algorithm is also used in the paper [8]. The EKF method has two stages: prediction and update. The prediction stage uses previous data estimations, and the update stage uses feedback correction to correct forecasted values. The data used here is to forecast the flood level. This algorithm's RMSE is 0.9236 m, according to this paper [8].

The main goal of this research is to predict floods more accurately using Binary Logistic Regression. It is possible to classify the flood dataset using binary classification. Fitting the model to the training dataset reveals which model parameters should be used to predict unknown labels on other data. So far, classic ML methods like SVC and KNN have shown better accuracy.

## II. PROPOSED METHOD

The purpose of this proposed paper is to answer if a higher accuracy rate using Binary Logistic Regression and lessening the error can be achieved. The dataset contains the record of the monthly rainfall index of Bangladesh, along with the yearly flood occurrence data near 34 stations. Sourced from the Bangladesh Meteorological Department

[9], this dataset will be used as input to make accurate predictions.

Firstly, the dataset containing the amount of rainfall and yearly flood occurrence of 34 stations in Bangladesh from the year 1980 to 2020 is organized and formatted to be fed to the system to preprocess. Secondly, the dataset has been categorized based on multiple parameters. Feature encoding and feature engineering need to be applied to preprocess the data as well. After partitioning the dataset in a 80:20 ratio for training and testing, the Binary Logistic Regression model has to be applied. Models like K-Nearest Neighbor (KNN), Support Vector Classifier (SVC) and Decision Tree Classifier then need to be applied for comparison. Lastly, it can be concluded which model is best to use to predict floods based on the accuracy and standard deviation of the models used.

### A. Dataset

The data was collected from Bangladesh Meteorological Department, Dhaka, Bangladesh, who are responsible for monitoring and issuing forecasts of all natural disasters to keep casualties to a minimum. They use rainfall data, satellite images and various other parameters to issue accurate weather forecasts.

Fig. 1. Rainfall Dataset

The dataset consists of the daily rainfall index and flood index of 34 stations in Bangladesh from 1980 to 2020. The index indicates how much precipitation is received corresponding to the long term average for a specified area and timeframe. In the rainfall dataset, the data has been recorded on a daily basis. The flood dataset has records of whether floods happened around the recorded stations from 1980-2020.

The dataset depending on multiple parameters has been categorized. The dataset will go through: data cleaning, feature engineering, feature encoding and feature scaling.

*Data Cleaning:* Firstly, in the dataset, unequal days of each month were handled. For example, some months have 31 days and some have 28 or 30 days. So, there was no data of rainfall for those specific days. Additionally, some days of particular stations did not have any data. Therefore, data imputation was applied to handle this issue. For instance, in the month of February,1980, the month ended at the 29th date, so for the 30th and 31st dates, the value zero for rainfall was added. Furthermore, in the dataset, the rainfall value for the 14th and 15th of 1983 November, Dhaka station, was missing, so, the value zero was also added here.

*Feature Engineering:* From the rainfall dataset, monthly rainfall data has been calculated and added into a new column. Then, the monthly rainfall data was set according to the particular stations and years. At this stage, the dataset contains the column of stations, years and all the 12 months.

After that, the flood data which was collected, was merged as a new feature into the rainfall data according to the stations and years.

*Feature Encoding:* The dataset that is used in this research paper has two attributes that have string type data which are - 'Station' and 'Flood'. As machine learning models give better results for numerical values, the string type data are encoded.

The attribute 'Station' contains all the names of the stations from where the daily rainfall data has been collected. Then by performing label encoding, the categorical values of the 'Station' column have been transformed into numerical values without adding any additional column.

In this dataset, the attribute 'Flood' has two unique values, 'YES' and 'NO' and to encode these values, binary encoding is used. After encoding the values of the feature 'Flood', the value 'YES' is replaced by 1 and the value 'NO' is replaced by 0.

Fig. 2. Dataset after feature engineering and feature encoding

*Feature Scaling:* Standard Scaler has been used on the dataset to make it unbiased and relevant to the models used. The data is scaled by centering them around the mean with a unit standard deviation. The formula for standardization can be defined as:

$$X = \frac{X - \mu}{\sigma}$$

where, $\mu$ is the mean of the feature values and $\sigma$ is the standard deviation of the feature values [10]. There are no restrictions on the range of the values. The dataset has been split into a train and test set with a ratio of 80:20. Then, the features have been scaled using the standard scaler.

## B. Machine Learning Models

Binary Logistic Regression, K-Nearest Neighbors (KNN), Support Vector Classifier (SVC), and Decision Tree Classifier (DTC) have all been used to predict values from the training set.

*Binary Logistic Regression:* The Binary Logistic Regression, like all other regressions analyses, is a predictive analysis that is used to describe data and explain the relationship between one dependent binary variable and one or more independent variables. The dependent variable has two categories, generally which are 1 for the occurrence of an event and 0 for its absence. A Logistic Regression can be interpreted as a specific case of generalized linear models with a dichotomous dependent variable [11]. A classical linear model can be denoted in the following manner:

$$Y = \alpha + \beta X + \varepsilon$$

where $Y$ is the dependent variable, $\alpha$ is the $Y$ intercept when $X$ is equal to zero, $X$ is the independent variable, $\beta$ is the regression coefficient representing the variation in $Y$ due to the change in values of $X$ and $\varepsilon$ is the error of the model. To categorize or limit the range of values for the dependent variable, the logistic regression model fits best. A graphical comparison between linear regression and logistic regression is shown below:

Replacing Y with a probability P that takes the range of probability to be within 0 and 1, the odds of P are taken,

$$\frac{P}{1-P} = \alpha + \beta X$$

In this equation, the range gets restricted which decreases the number of datapoints, eventually decreasing the correlation. To avoid this, the log of the odds need to be taken and exponent has to be added to both sides and the solution for $P$ is:

$$P = \frac{1}{1 + e^{-(\alpha + \beta X)}}$$

This is the sigmoid function for the logistic regression model used to predict any dichotomous dependent variable [12]. According to the dataset, the independent variable is the amount of annual rainfall and the dependent variable remains whether there will be a flood based on the rainfall or not.

*Support Vector Classifier:* The Support Vector Classifier (SVC) is a machine learning algorithm that uses both regression and classification. Structural Risk Minimization Principle is roughly implemented in this way. The SVC, in contrast to other models, aims to fit the best line within a threshold value rather than to reduce error between real and predicted values. According to this research article [13], a training SVC algorithm creates a model which designates the data to one class or another developing a binary and non probabilistic linear classifier. These classes are clearly separated from each other with the help of a gap or a spatial line.

SVC predicts the newer data and the class they belong to considering the distance of these classes from the line. The basic idea is to perform linear regression to find a decision function for a given sample x,

$$\sum_{i \in SV} y_i \alpha_i K(x_i, x) + b$$

The term $\alpha_i$ are called the dual coefficients which are upper-bound by C and $b$ is an independent term that has to be estimated. $K(x_i, x)$ is the kernel where, $x$ is the input vector [9].

*K-Nearest Neighbor:* The KNN algorithm is a sort of supervised machine learning method that is being used to solve both classification and regression predicting problems. The KNN approach uses feature similarity to forecast the values of new data points, which means that the new data point will be assigned a value based on how closely it resembles the points in the training set. According to this research paper [13], the K nearest neighbor (KNN) is a non-parametric algorithm that can be used for regression predictive problems. The KNN method assumes the resemblance between the new data and the existing data and places the new data in the category that is most comparable to the existing categories. The KNN algorithm calculates the distance between a new data point and all previous data points in the training set. There are a variety of distance functions for calculating the distance but Euclidean is the most widely utilized method.

Euclidean distance:

$$d(x, y) = \sqrt{\sum_{i=1}^{m} (x_i - y_i)^2}$$

*Decision Tree Classifier:* Decision Trees Classifier (DTC) are mainly used for regression and classification which is a non-parametric supervised learning method. By using decision trees, models can be created which will be able to predict target variables through learning simple decision rules deduced from the features of the dataset [14]. Decision trees help to explain the decisions of predictive models graphically and in decision trees the internal nodes denote the test on the features, branches denote the outcome and the leaf nodes denote the final decision that is derived from computing the features [15]. One of the most important factors of decision trees is to create a sequence of splits and for splitting it separates the data into two groups that are the purest. For calculating the purity of groups, decision trees calculate entropies of those groups. Entropy of a decision tree with C classes:

Entropy:

$$\sum_{i}^{c} - p_i \log_2 p_i$$

Here, $p_i$ means the probability of randomly picking an element of class $i$. The entropy values are ranged between 0 to 1, where 1 means maximum impure groups and 0 means full pure groups. Another statistical property of decision trees is information gain which represents decrease in entropy. It calculates the difference between the dataset's entropy before and after splitting depending on specified feature values.

InformationGain:

$$Entropy(before) - \sum_{j-1}^{k} Entropy(j, after)$$

Here, "$before$" refers to the dataset prior to the split, $k$ refers to the number of subsets formed by the split, and $(j, after)$ refers to subset $j$ following the split [16].

III. RESULTS

For this research, we have used Binary Logistic Regression, Support Vector Classifier (SVC), K-Nearest Neighbor (KNN) and Decision Tree Classifier (DTC) for predicting our data. The data has been split into 80:20 ratio for training and testing the models. The classification has been done on 16 columns: Station, Year, 12 months and Flood index. Firstly the models have been implemented on the whole dataset which consists of data from 1980-2020. Later, the same models have been implemented with a shorter timeline of 10 years, 2011-2020 to check the accuracy and compare with the previous implementation.

A. Timeline: 1980 to 2020

TABLE I. RESULTS OF IMPLEMENTED MODELS (TIMELINE: 1980-2020)

| Machine Learning Models | Accuracy | Precision | Precision |
|---|---|---|---|
| Binary Logistic Regression | 0.8561 | 0.75 | 0.55 |
| Support Vector Classifier (SVC) | 0.8409 | 0.7647 | 0.4333 |
| K-Nearest Neighbors (KNN) | 0.8371 | 0.7576 | 0.4167 |
| Decision Tree Classifier (DTC) | 0.7879 | 0.5303 | 0.5833 |

From the table, Binary Logistic Regression has the highest accuracy rate of 0.8561 with a precision and recall score of 0.75 and 0.55 respectively.

```
              precision    recall  f1-score   support

           0       0.88      0.95      0.91       204
           1       0.75      0.55      0.63        60

    accuracy                           0.86       264
   macro avg       0.81      0.75      0.77       264
weighted avg       0.85      0.86      0.85       264
```

Fig. 4. Classification Report of Binary Logistic Regression (1980-2020)

After that, the Support Vector Classifier (SVC) has the highest accuracy of 0.8409 with a precision of 0.7647 which is higher than the Binary Logistic Regression.

```
              precision    recall  f1-score   support

           0       0.85      0.96      0.90       204
           1       0.76      0.43      0.55        60

    accuracy                           0.84       264
   macro avg       0.81      0.70      0.73       264
weighted avg       0.83      0.84      0.82       264
```

Fig. 4. Classification Report of Support Vector Classifier (SVC) (1980-2020)

Then, K-Nearest Neighbors (KNN) has an accuracy of 0.8371, precision and recall score respectively 0.7576 and 0.4167.

```
              precision    recall  f1-score   support

           0       0.85      0.96      0.90       204
           1       0.76      0.42      0.54        60

    accuracy                           0.84       264
   macro avg       0.80      0.69      0.72       264
weighted avg       0.83      0.84      0.82       264
```

Fig. 5. Classification Report of K-Nearest Neighbors (KNN) (1980-2020)

Lastly, Decision Tree Classifier (DTC) has an accuracy of 0.7879 which is the lowest among the models that have been used but it has the highest recall score of 0.5833.

```
              precision    recall  f1-score   support

           0       0.87      0.85      0.86       204
           1       0.53      0.58      0.56        60

    accuracy                           0.79       264
   macro avg       0.70      0.72      0.71       264
weighted avg       0.80      0.79      0.79       264
```

Fig. 6. Classification Report of Decision Tree Classifier (DTC) (1980-2020)

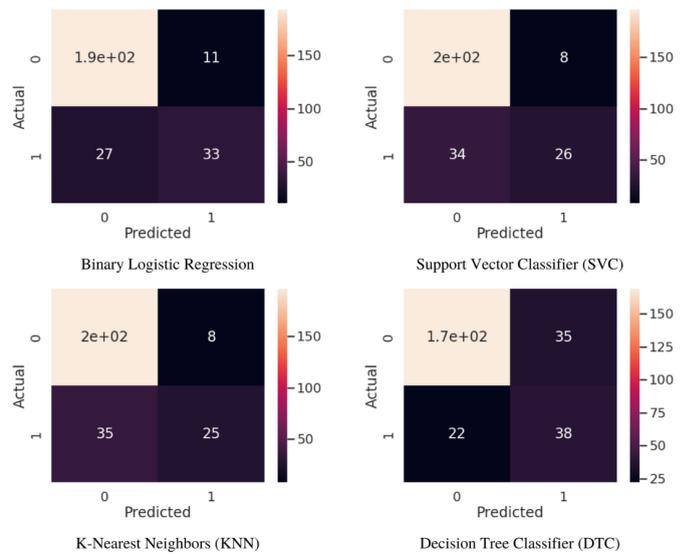

Binary Logistic Regression    Support Vector Classifier (SVC)

K-Nearest Neighbors (KNN)    Decision Tree Classifier (DTC)

Fig. 7. Confusion matrix of the used models (1980-2020)

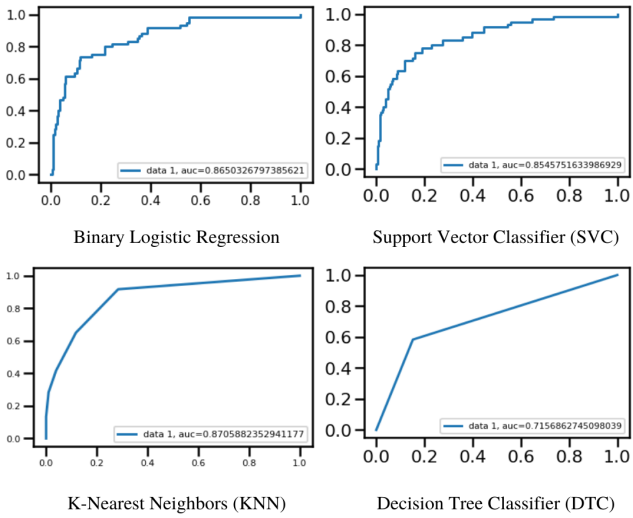

Fig. 8. ROC Curves of used models (2011-2020)

B. *Timeline: 1980 to 2020*

TABLE II. RESULTS OF IMPLEMENTED MODELS (TIMELINE: 2011-2020)

| Machine Learning Models | Accuracy | Precision | Precision |
|---|---|---|---|
| Binary Logistic Regression | 0.8676 | 0.6154 | 0.667 |
| Support Vector Classifier (SVC) | 0.8088 | 0.4667 | 0.5833 |
| K-Nearest Neighbors (KNN) | 0.8235 | 0.50 | 0.50 |
| Decision Tree Classifier (DTC) | 0.8088 | 0.4545 | 0.4167 |

From the table, Binary Logistic Regression has the highest accuracy rate of 0.8676 with a highest precision and also highest recall score of 0.6154 and 0.6667 respectively.

```
              precision    recall  f1-score   support

           0       0.93      0.91      0.92        56
           1       0.62      0.67      0.64        12

    accuracy                           0.87        68
   macro avg       0.77      0.79      0.78        68
weighted avg       0.87      0.87      0.87        68
```

Fig. 9. Classification Report of Binary Logistic Regression (2011-2020)

After that, the Support Vector Classifier (SVC) has accuracy of 0.8088 with a precision of 0.4667 and recall score of 0.5833.

```
              precision    recall  f1-score   support

           0       0.91      0.86      0.88        56
           1       0.47      0.58      0.52        12

    accuracy                           0.81        68
   macro avg       0.69      0.72      0.70        68
weighted avg       0.83      0.81      0.82        68
```

Fig. 10. Classification Report of Support Vector Classifier (SVC) (2011-2020)

Then, K-Nearest Neighbors (KNN) has an accuracy of 0.8235 which is the highest after Binary Logistic Regression, with equal precision and recall score of 0.50.

```
              precision    recall  f1-score   support

           0       0.89      0.89      0.89        56
           1       0.50      0.50      0.50        12

    accuracy                           0.82        68
   macro avg       0.70      0.70      0.70        68
weighted avg       0.82      0.82      0.82        68
```

Fig. 11. Classification Report of K-Nearest Neighbors (KNN) (2011-2020)

Lastly, Decision Tree Classifier (DTC) has an accuracy of 0.8088 which is equal to Support Vector Classifier (SVC), respectively precision and recall score of 0.4615 and 0.4167.

```
              precision    recall  f1-score   support

           0       0.89      0.88      0.88        56
           1       0.46      0.50      0.48        12

    accuracy                           0.81        68
   macro avg       0.68      0.69      0.68        68
weighted avg       0.82      0.81      0.81        68
```

Fig. 12. Classification Report of Decision Tree Classifier (DTC) (2011-2020)

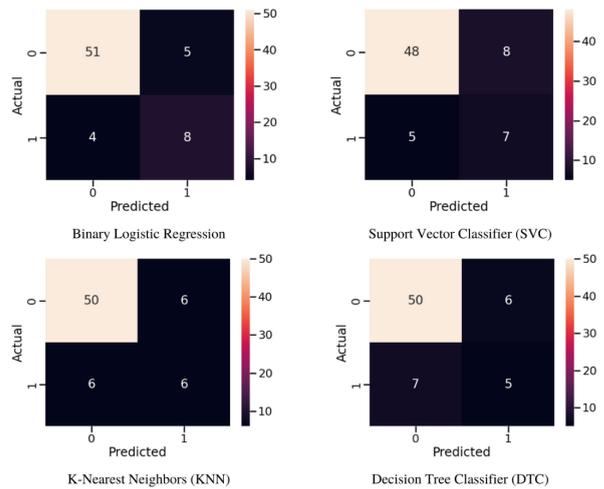

Fig. 13. Confusion matrix of the used models (2011-2020)

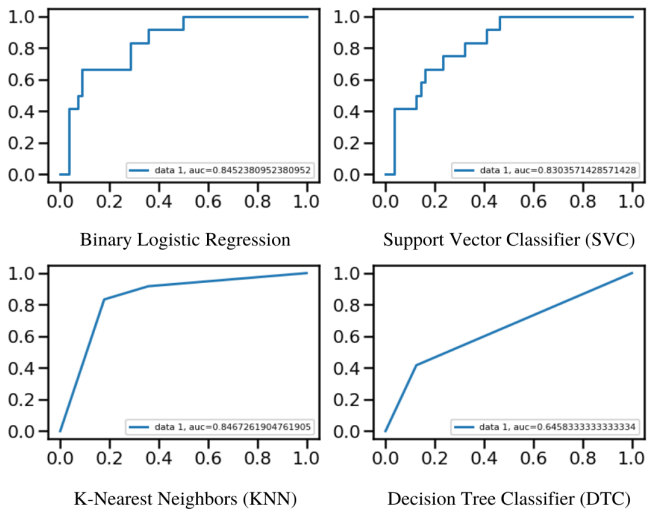

Fig. 14. ROC Curves of used models (2011-2020)

## IV. Discussion

For the timeline 1980-2020, from the ROC Curves of Fig. 10, it is shown that the K-Nearest Neighbors (KNN) model has the highest AUC score of 0.87 but from Table 1 and figure 13, K-Nearest Neighbors (KNN) model has less accuracy and f1 score than Binary Logistic Regression Model (Accuracy: 0.8561, f1 score: 0.86). As higher the f1 score the better, it can be stated that Binary Logistic Regression is better than any other model with a better accuracy and other scores.

For the timeline 2011-2020, from the ROC Curves of Fig. 17, it is shown that the Binary Logistic Regression model and the K-Nearest Neighbors (KNN) model have almost AUC scores (0.845 and 0.846 respectively). As the Binary Logistic Regression model has a better accuracy (0.8676) and f1 score (0.87), this model is considered to be the better one among the four models.

From both of the timelines, most of the models gave better accuracy on the 10 years of rainfall data (2011-2020) than the whole timeline (1980-2020). Binary Logistic Regression has given the highest accuracy of 86.76% (Timeline: 2011-2020) among all with a better accuracy, recall and f1-score.

## V. Conclusion

As climate changes over the years and depending on other parameters, the thresholds for floods are changing. That's why the shorter timeline of data gives slightly better accuracy. Since these change over a long time period, in this research, the models gave higher accuracy with a shorter time range. Also, due to time constraint only the rainfall data along with flood occurrence was manageable. There are more factors related to flood like, river water level, temperature, humidity, other natural disasters etc. In the future, this research paper would attempt to develop the models further by adding the other factors and correlating them.